\documentclass[letterpaper, 10 pt, conference]{ieeeconf}  % Comment this line out if you need a4paper

\IEEEoverridecommandlockouts                              % This command is only needed if 
\usepackage{amsmath}
\usepackage{amssymb}
\usepackage{bm}
\usepackage{booktabs}
\usepackage{cite}
\usepackage{graphicx}
\usepackage{lettrine}
\usepackage{multirow}
\usepackage{subcaption}
\usepackage[table]{xcolor}
\usepackage{xspace}

\usepackage{hyperref}
\newcommand{\method}{\textit{OmniV2X}\xspace}

\title{
\LARGE{\textbf{%
    OmniV2X: A Generative Foundation Planner for\\
    Efficient End-to-End Cooperative Driving
}}
}

\author{
Juntong Peng, Juanwu Lu, Yupeng Zhou, Can Cui, Yaobin Chen, Ziran Wang
\thanks{The authors are with the College of Engineering, Purdue University, West Lafayette, IN 47907, USA.
Corresponding author: Juntong Peng~{\tt\small juntong@purdue.edu}.}%
}

\begin{document}

\maketitle
\thispagestyle{empty}
\pagestyle{empty}

%%%%%%%%%%%%%%%%%%%%%%%%%%%%%%%%%%%%%%%%%%%%%%%%%%%%%%%%%%%%%%%%%%%%%%%%%%%%%%%%
% Main sections
\begin{abstract}

We present~\method, a generative foundation model for vehicle-to-everything (V2X) cooperative driving. The model directly interprets independent context sequences comprising multi-modal and multi-agent observations. The new design mitigates the computational cost of dense 3D perception, the vulnerability to data scarcity in cooperative scenarios, and the poor compliance with standardized messaging in existing methods that fuse multi-modal inputs into a shared representation. For training, we present an end-to-end supervised pipeline using a downstream trajectory generation loss, in which a high-capacity generative sequence planner implicitly learns to steer the model and leverage multi-modal inputs via cross-attention injection. As a foundation model, we demonstrate that~\method pre-trained on large-scale single-agent planning datasets can efficiently adapt to cooperative environments by integrating the conditioning context with lightweight, standard-compliant V2X tokens. Evaluated on the DAIR-V2X-Seq dataset,~\method outperforms existing end-to-end cooperative driving baselines, achieving state-of-the-art performance with less than $10\%$ of the fine-tune V2X dataset and less than $1\%$ of the communication bandwidth. We conduct comprehensive evaluations to demonstrate its computational efficiency and robustness under real-world constraints. The code and model checkpoints are available at:
\url{https://github.com/JuntongPeng/OmniV2X}.

\end{abstract}

\section{INTRODUCTION}
\label{sec: intro}

% Paragraph 1: The Promise & The Pioneers
End-to-end (E2E) autonomous driving is evolving rapidly, yet single-vehicle systems remain fundamentally constrained by ego-centric perception. Cooperative driving with vehicle-to-everything (V2X) communications promises to overcome physical occlusions in single-vehicle perception by extending the perception range through the sharing of infrastructure perspectives. Pioneering works such as UniV2X\cite{yu_end--end_2025} and UniMM-V2X\cite{song_unimm-v2x_2025} demonstrate the benefits of E2E cooperative driving by feature-based fusion at both the perception and prediction levels.
However, real-world deployments of these methods encounter several systemic bottlenecks. Transmitting bird's-eye view (BEV) features requires substantial communication bandwidth, and the unique feature spaces across different models hinder standardization and cross-model sharing among connected and automated vehicles (CAVs). Furthermore, aligning multi-agent features onto a shared spatial grid tightly couples the ego and infrastructure pipelines, requiring exhaustive 3D bounding-box supervision on temporally synchronized datasets. While there is an increasing number of perception-label-free ego-centric driving datasets and benchmarks in both scale and diversity\cite{caesar_nuscenes_2020, dauner_navsim_2024}, real-world cooperative datasets\cite{yu_v2x-seq_2023} remain limited due to the high cost of equipping infrastructure and synchronizing across sources. Therefore, it requires a disruptive paradigm shift to enable the effective use of multi-modal V2X inputs.

\begin{figure}[t]
    \centering
    \includegraphics[width=\columnwidth]{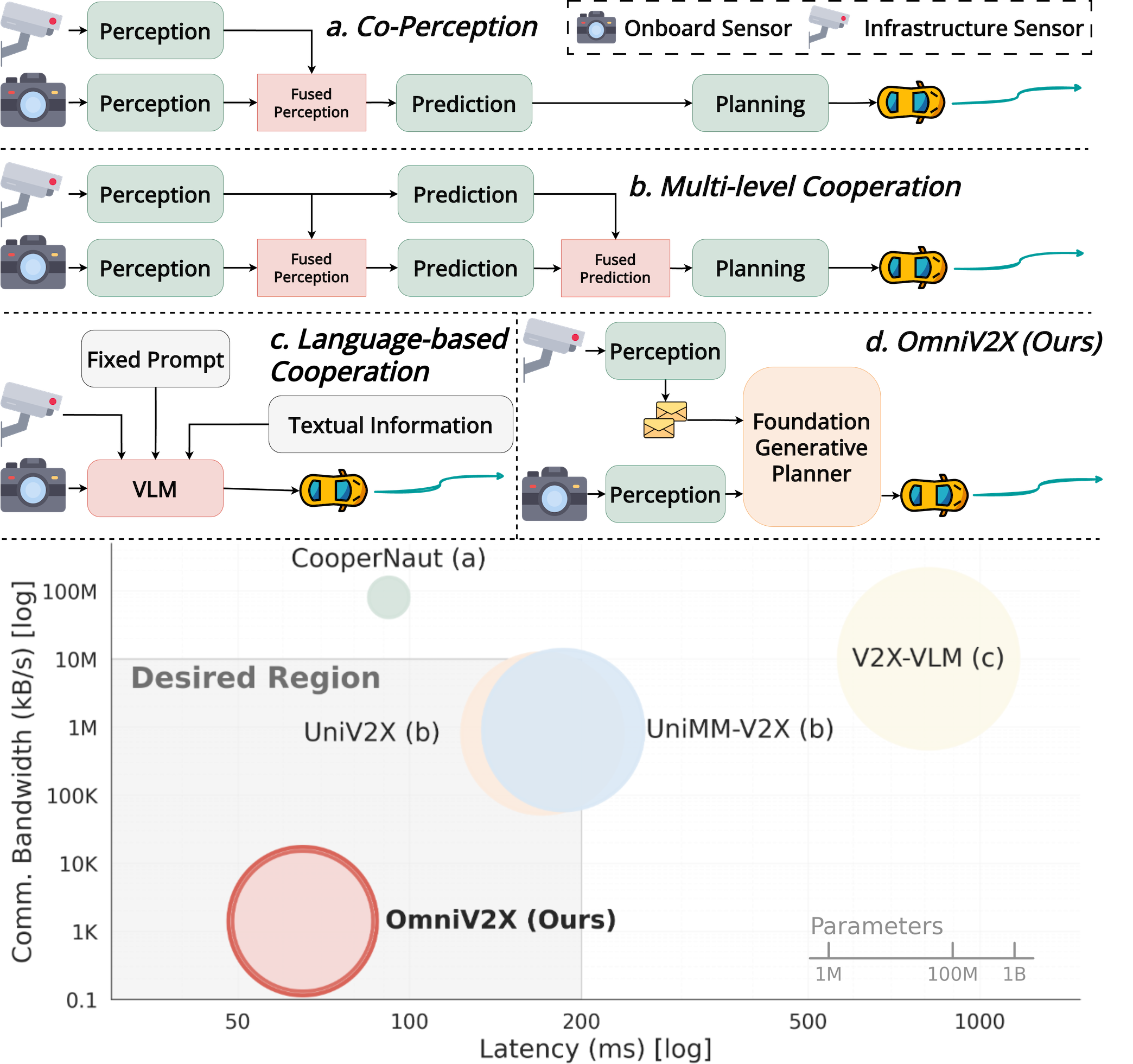}
    \caption{\textit{Top:} Comparison among different paradigm in end-to-end cooperative driving.
    %Co-Perception and Multi-level Cooperation demonstrate superior planning performance, but require dedicated design fusion modules and force a spatially shared feature space. Language-based Cooperation leverages the reasoning capabilities of language models, but its computational complexity makes it unsuitable for safety-critical scenarios. The chart at the bottom showcases the deployment-relevant performance: 
    \textit{Bottom:} \method successfully provides \textbf{low latency, low communication bandwidth, and a reasonable number of parameters.}}
    \label{fig:teaser}
    \vspace{-2em}
\end{figure}

To address these limitations, we propose shifting the focus of cooperative driving from rigid spatial feature fusion to conditional generative sequence modeling. We introduce~\method, a scalable framework for E2E Cooperative Driving. \method treats the ego-centric sensor input, potential infrastructure-sourced information, and other auxiliary observations as condition tokens for future waypoint generation. Rather than explicitly merging features from different sources into a shared representation,~\method presents heterogeneous, multi-agent inputs to the generative planner as independent context sequences. Primarily supervised by a downstream trajectory-generation loss, the generative planner implicitly learns to extract and utilize these distinct modalities on its own. Pre-trained on large-scale ego-only datasets, the learned generative driving prior can be readily adapted to cooperative inputs and novel environments with only a minimal amount of target data. Our main contributions are as follows:
\begin{itemize}

    \item \textbf{A Scalable V2X Cooperative Driving Framework.} \method bypasses rigid spatial-temporal fusion of multi-agent perceptual features, learning directly from the downstream planning task. This design treats heterogeneous inputs as independent context sequences, enabling seamless integration of SAE-standard-compliant V2X messages.
    
    \item \textbf{Data-Efficient Ego-to-V2X Knowledge Transfer.} \method bridges the real-world cooperative data gap by leveraging a domain-stable feature space to transfer knowledge from single-agent driving. The pre-trained model achieves significant optimization acceleration, surpassing from-scratch multi-agent baselines using less than 10\% of the target V2X data.

    \item \textbf{Inference-Efficient End-to-end Cooperative Driving.} The standardized V2X messages only consume communication bandwidth less than \textit{1.4kBPS}. Moreover, with the design of fully BEV-free processing, \method removes expensive feature encoding on dense spatial grids, thereby becoming highly efficient and requiring only \textit{550 MB} GPU memory during inference, with an inference latency of \textit{35 ms}. 
    
    \item \textbf{Comprehensive Evaluation \& Field Testing.} \method outperforms heavily-supervised fusion architectures at real-time speeds, while requiring less than 1\% of the communication bandwidth of current baselines on the DAIR-V2X dataset. Extensive ablation studies systematically assess how pretraining benefits cooperative driving. We also validate the performance of~\method against real-world challenges through various noise-injection tests and V2X-augmented field tests.
\end{itemize}
% --- FIGURE 1: TEASER ---

%%%%%%%%%%%%%%%%%%%%%%%%%%%%%%%%%%%%%%%%%%%%%%%%%%%%%%%%%%%%%%%%%%%%%%%%%%%%%%%%
\section{RELATED WORK}
\label{sec:related_work}

\subsection{End-to-End Autonomous Driving}
The paradigm of autonomous driving is evolving from modular systems, which suffer from compounding errors, to E2E architectures that directly map sensor inputs to planning trajectories or control commands. Pioneering works such as UniAD \cite{hu_planning-oriented_2023} demonstrated that jointly optimizing perception and planning in a shared Bird's-Eye View (BEV) space significantly improves planning performance. Moreover, VAD \cite{jiang_vad_2023}, Hydra-MDP \cite{li_hydra-mdp_2024}, and SparseDrive \cite{sun_sparsedrive_2025} showed that sparse instance-level or query-based representations could reduce the computational overhead of dense BEV grids. Extending beyond deterministic heads, recent approaches, such as GenAD \cite{zheng_genad_2024} and DiffusionDrive \cite{liao_diffusiondrive_2025}, leverage continuous generative models to refine target waypoints from anchored trajectory candidates. However, all these single-agent systems remain bounded by the physical limitations of ego-centric perception. In complex urban environments, onboard cameras are heavily constrained by line-of-sight occlusions.

\subsection{Language Foundation Models for Autonomous Driving}
Driven by the success of large language models (LLMs) and vision-language models (VLMs), recent works increasingly adapt language foundation models for E2E driving \cite{sima_drivelm_2024,wang_drivecot_2024,hwang_emma_2024}. Leveraging their broad reasoning capabilities, methods such as Alpamayo-R1 \cite{nvidia_alpamayo-r1_2026} and ReCogDrive \cite{li_recogdrive_2025} successfully integrate semantic understanding with continuous trajectory generation.

Deploying these high-capacity models on resource-constrained vehicles, however, remains a critical challenge. While techniques like masked decoding \cite{cui_vilad_2025} and structured chain-of-thought \cite{gu_accelerating_2026} accelerate discrete text decoding, adapting these language-native architectures to driving exposes three structural bottlenecks: the massive compute overhead of billion-parameter weights, a representational mismatch between discrete tokens and continuous kinematics, and prohibitive memory scaling when processing increasing contexts. Consequently, real-time autonomous driving requires shifting from language-centric architectures to a \textit{generative foundation planner} natively optimized for continuous spatial sequences.

\subsection{Cooperative Driving with V2X Communications}

V2X communications mitigate the physical occlusion limits of ego-centric perception. Early systems established the paradigm of intermediate feature fusion, projecting multi-agent data into shared spatial grids for co-perception and cooperative planning \cite{vedaldi_v2vnet_2020,cui_coopernaut_2022,xu_cobevt_2023}. Recently, E2E architectures such as UniV2X \cite{yu_end--end_2025} and UniMM-V2X \cite{song_unimm-v2x_2025} have successfully extended the BEV fusion paradigm to tracking, prediction, and trajectory planning. However, existing feature-based methods require rigid alignment of feature spaces among collaborators. This limitation significantly hinders the deployment of V2X cooperative driving in real-world scenarios. Recent methods such as V2X-VLM \cite{you_v2x-vlm_2025}, Lang-Coop \cite{gao_langcoop_2025}, and CoLMDriver \cite{liu_colmdriver_2025} use natural language as the universal message. While these language-based methods are beneficial for long-term decision-making, the latency incurred by multi-round LLM inference makes them impractical for safety-critical short-term planning.

To address the alignment challenge of feature-based fusion and the inference latency of language models, \method formulates cooperative driving as a continuous conditional sequence generation task that incorporates context information from multiple sources. By representing heterogeneous multi-agent inputs as independent context sequences and processing them via cross-attention injection, our generative foundation planner can consume SAE-compliant V2X messages and overcome quadratic memory bottlenecks, thereby enabling real-time, safety-critical V2X autonomy.

\section{METHODOLOGY}
\label{sec:methodology}

\begin{figure*}[t]
    \centering
    \includegraphics[width=0.95\linewidth]{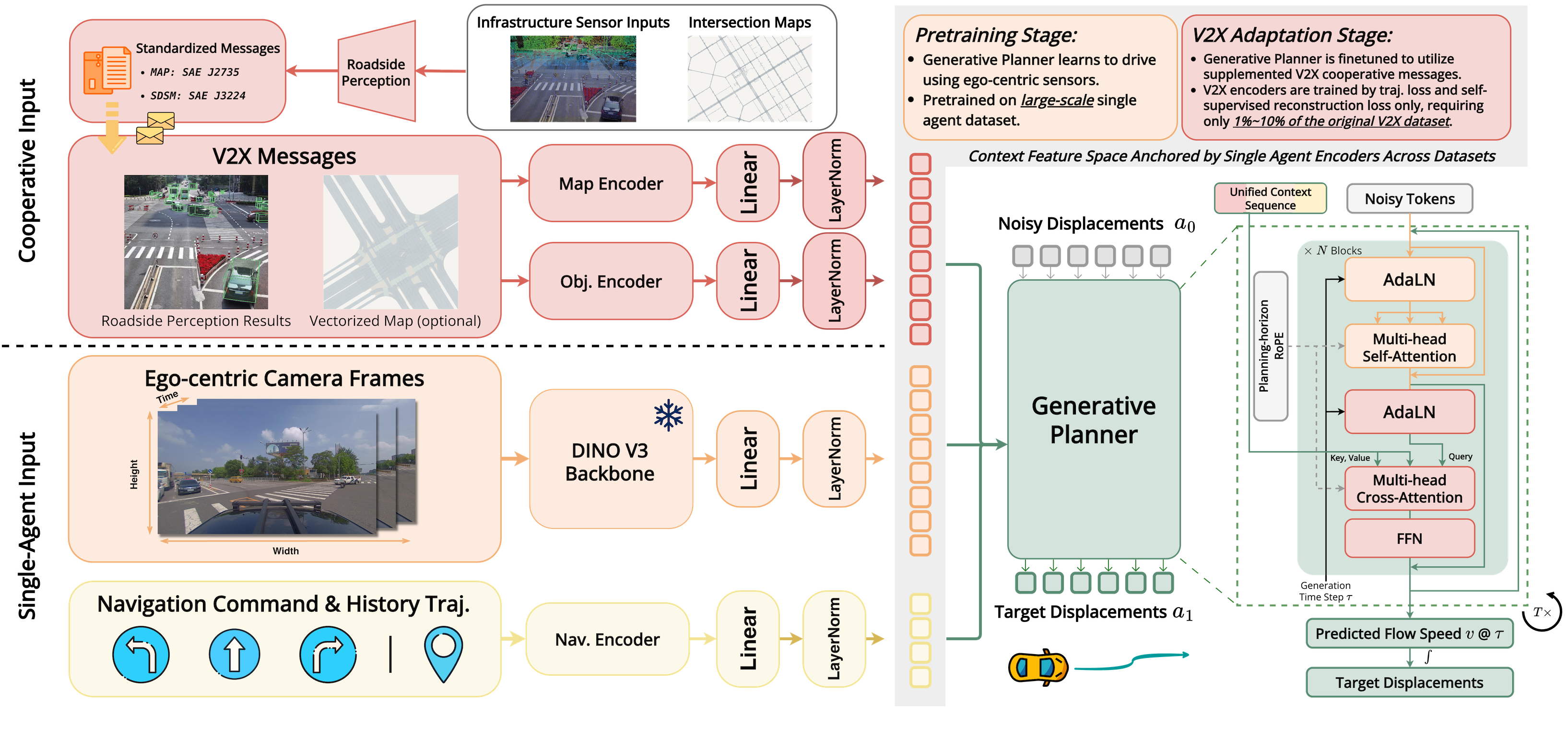}
    \caption{The \method architecture. All modalities are independently encoded and normalized, then concatenated into a unified context sequence, which is consumed by a generative planner via computationally efficient cross-attention. The model is first pretrained on a large-scale single-agent dataset and adapted to a smaller-scale V2X dataset.}
    \label{fig: architecture}
    \vspace{-1.5em}
\end{figure*}

% --- OPENING REMARK (THE PREAMBLE) ---
This section introduces the formulation and architectural design of~\method. We reframe E2E cooperative driving as a multi-input conditional sequence generation problem, shifting away from recent methods extended from BEV-based perception\cite{yu_end--end_2025,song_unimm-v2x_2025,zhu_m3cad_2025,zhu_features_2026}. As illustrated in Fig.~\ref{fig: architecture}, our framework achieves this through three primary components: (1)~independent modules transforming heterogeneous inputs, including ego-centric visual features, infrastructure V2X messages, and high-level navigational commands into the unified conditioning sequence consumed by the planner; (2)~a generative foundation model serving as the core planning engine, optimized for the long-context conditions from different sources via cross-attention injection; and (3)~a perception-label-free training pipeline including ego-centric driving pretraining and V2X cooperative driving fine-tuning. We detail these components in the following.

\subsection{Independent Modules for Context Tokenization}
\label{sec:meth_tokens}
To construct the unified context sequence $\bm{C}$ required by the conditional generative planner, \method encodes each input modality through an independent neural network module to a unified $D$-dimensional hidden space.

\subsubsection{Projection-Free Visual Context}

A critical computational bottleneck in intermediate BEV fusion is the reliance on rigid perspective projections (\textit{e.g., view transformers}) to explicitly map 2D image pixels to 3D coordinates. Although this projection benefits precise 3D perception, tracking, and prediction, it incurs substantial computational cost in dense 3D space. As an alternative, our generative planner learns to implicitly map the latent 2D visual sequence directly to continuous kinematic displacements. We extract visual context from the front-view camera using a pretrained vision backbone as the primary condition for future waypoint planning. 

Crucially, to ensure a stable feature space across multiple datasets, the backbone is kept \textbf{completely frozen} throughout training and serves as a general-purpose visual feature extractor. A single trainable linear projection layer then extracts visual patch tokens directly to the hidden dimension $D$, forming the front-view visual context $\bm{C}_{\text{FV}}$. The downstream generative planner, as we will describe in Sec.~\ref{sec:gfm}, will learn to use these general-purpose image features to yield safe and efficient planning results entirely from minimizing the waypoints loss. The visual patch tokens from two consecutive frames are concatenated along the sequence length dimension to provide implicit temporal motion cues.

\subsubsection{High-level Navigation Context}
To define the ego-agent's intent, an independent module constructs the token sequence $\bm{C}_{\text{nav}}$. We encode high-level navigation commands (\textit{e.g., Turn Left, Go Straight}) through learned embeddings. The structure of navigation commands follows the practice of previous work\cite{song_unimm-v2x_2025}. Notably, ego-status, such as current speed, throttle, and steering, is excluded to prevent the generative planner from exploiting kinematic modelling as a shortcut, in which the model learns a trivial policy of simply extrapolating its current velocity rather than actively reasoning about the dynamic visual and cooperative environment.

\subsubsection{SAE-Compliant V2X Infrastructure Context}

To facilitate the use of real-world communication standards in E2E cooperative driving, infrastructure data is treated as instance-level context. While \method eliminates the need for rigid camera calibration to build a BEV representation, multi-agent cooperation still necessitates basic spatial alignment. Herein, we assume the Roadside Unit (RSU) relies on an arbitrary detector to broadcast detected objects in a global coordinate system. The ego-vehicle utilizes its onboard localization (GNSS) to transform these states into its local coordinate frame.

For each infrastructure-detected object, a $13$-dimensional state vector is decoded from the Sensor Data Sharing Message (SDSM) following the definition in SAE-J3224\cite{SAEJ3224_2022}:
\begin{equation}
    \bm{v}_{\text{obj}} = [x, y, z, l, w, h, \sin\theta, \cos\theta, v_x, v_y, c_{\text{veh}}, c_{\text{ped}}, c_{\text{cyc}}]
\end{equation}
where $c$ represents one-hot class encodings. The orientation is encoded continuously via $[\sin\theta, \cos\theta]$ to prevent angular wrapping discontinuities. Furthermore, optional map information can also be decoded similarly from the MAP message~\cite{SAEJ2735_2024}. These vectors, combining infrastructure-detected objects and maps, are encoded by a lightweight Transformer to form the infrastructure context $\bm{C}_{\text{v2x}}$.

\subsubsection{Context Alignment and Sequence Concatenation}
Because visual patches and infrastructure-sourced bounding boxes possess extremely different numerical scales, concatenating them directly can lead to modality collapse. To prevent visual features from dominating others, each modality is independently normalized using Layer Normalization (LN). A learnable additive Modality Embedding is added to distinguish different data sources. The final conditioning sequence $\bm{C}$ is formally constructed via sequence concatenation:
\begin{equation}
\resizebox{\linewidth}{!}{$
\bm{C} =
\bigl[\text{LN}(\bm{C}_{\text{FV}}) + \bm{S}_{\text{FV}}
\parallel \text{LN}(\bm{C}_{\text{nav}}) + \bm{S}_{\text{nav}}
\parallel \text{LN}(\bm{C}_{\text{v2x}}) + \bm{S}_{\text{v2x}}
\bigr]
$}
\end{equation}
(where $\parallel$ denotes sequence concatenation along the sequence length dimension and $S_{*}$ denotes an additive learnable segment embedding.)

\subsection{Generative Foundation Model for Efficient Conditional Waypoint Generation}
\label{sec:gfm}

Following the first architectural component, we formulate cooperative trajectory planning as a conditional generative sequence task. While our tokenized interface by design supports various sequence backbones, including one-shot, autoregressive, and iterative-refinement decoders, we build our primary foundation model using a Rectified Flow Transformer.

\subsubsection{Continuous Displacement Generation via Rectified Flow}
Rather than regressing absolute coordinates (even within the ego-centric coordinate frame), which introduces spurious spatial correlations between static context information and the generation outputs, we use the step-wise positional displacements as generation targets. We find the design crucial to avoid the potential shortcut learning of simply memorizing routes at specific physical intersections during the cooperative fine-tuning stage, especially when vectorized maps exist. We observe that, under an identical hyperparameter, using waypoints as the generation target leads to a severe over-fitting issue, and the best L2 error is 57\% worse than the displacements generation counterpart.

During training, we define a vector field $v_\theta(\bm{X}_\tau, \tau, \bm{C})$ directing noisy samples $\bm{X}_\tau$ back to the ground-truth trajectory $\bm{X}_1$, where $\bm{X}_\tau$ is a linearly interpolated state $\bm{X}_\tau = \tau\bm{X}_1+(1-\tau)\bm{X}_0$. The model is optimized using a standard Flow Matching Mean Squared Error (MSE) objective:
\begin{equation}
    \mathcal{L}_{\text{flow}} = \mathbb{E}_{\tau, \bm{X}_0, \bm{X}_1} \left[ \| v_\theta(\bm{X}_\tau, \tau, \bm{C}) - (\bm{X}_1 - \bm{X}_0) \|^2_2 \right]
\end{equation}
Furthermore, Rectified Flow yields straight-path probability flows, allowing flexible inference-time computation. We train the ODE integration with high-fidelity steps, but dynamically truncate waypoint generation to fewer evaluation steps during deployment, striking an optimal balance between generation quality and efficiency.

Notably, during the generation, as there is no causal mask to distinguish between temporal steps $\Delta \bm{x}_1$ and $\Delta \bm{x}_t$, we apply rotary position embedding\cite{su_roformer_2024} on the noisy displacement tokens to provide a strong temporal signal.

\subsubsection{Complexity-Optimized Conditioning via Cross-Attention Injection}

An important design choice in \method is how the multi-modal context sequence $\bm{C}$ conditions the generated trajectory. \textit{Prefix conditioning} has proven effective in language foundation models, in which the generation context is prefilled and an autoregressive decoder iteratively generates the next token. This approach concatenates the condition sequence of length $L$ with the generated target horizon of length $H_t$ at time step $t$ and then computes causal self-attention over the joint sequence. This attention calculation yields a computational complexity of $\mathcal{O}((L+H)^2) = \mathcal{O}(L^2 + 2LH + H^2)$. 

In the context of E2E cooperative driving, the context length (comprising hundreds of visual patches and V2X tokens) vastly exceeds the future waypoint horizon ($L \gg H$). Consequently, the $\mathcal{O}(L^2)$ term becomes a critical computational bottleneck, especially regarding memory requirements. To resolve this, \method utilizes \textit{cross-attention injection}. The generated waypoint tokens perform self-attention in $\mathcal{O}(H^2)$ to maintain temporal kinematic consistency, and query the context sequence $\bm{C}$ exclusively via cross-attention in $\mathcal{O}(LH)$. Unlike the recent query-based BEV-fusion paradigm, our action-sourced queries ($\bm{Q}$) serve as active probes, retrieving routing and safety-related features from conditioning tokens. By removing the inter-context self-attention overhead among the condition tokens, the attention complexity is drastically reduced to $\mathcal{O}(LH + H^2)$, enabling real-time generation on edge computing hardware.

\begin{table*}[t]
\centering
\caption{Quantitative evaluation of open-loop trajectory planning on the DAIR-V2X cooperative benchmark. Results for \method are reported as the mean $\pm$ standard deviation across 10 different seeds. Baseline results are reported by UniMM-V2X~\cite{song_unimm-v2x_2025}. Methods in \textcolor{gray}{gray text} utilize vector maps. Best comparable results are in \textbf{bold}.}
\label{tab:main_results}

\footnotesize
\resizebox{\textwidth}{!}{
\begin{tabular}{@{} c l l c cccc c c @{}}
\toprule

& \multirow{2}{*}{\textbf{Method}} &
\multirow{2}{*}{\textbf{\begin{tabular}[c]{@{}c@{}}V2X Message\\Type\end{tabular}}} &
\multirow{2}{*}{\textbf{\begin{tabular}[c]{@{}c@{}}Trans. Cost\\ (BPS) $\downarrow$\end{tabular}}} &
\multicolumn{4}{c}{\textbf{L2 Error (m) $\downarrow$}} &
\multirow{2}{*}{\textbf{\begin{tabular}[c]{@{}c@{}}3s Avg. Col.\\ Rate (\%) $\downarrow$\end{tabular}}} &
\multirow{2}{*}{\textbf{PDMS $\uparrow$}} \\
\cmidrule(lr){5-8}

& & & & \textbf{1s} & \textbf{2s} & \textbf{3s} & \textbf{Avg.} & & \\
\midrule

% ROTATED GROUP 1: Single-Agent
\multirow{3}{*}{\rotatebox[origin=c]{90}{\scriptsize \textit{Ego-Only}}}
& VAD \cite{jiang_vad_2023} \textcolor{gray}{\scriptsize [ICCV'23]}                & \multicolumn{1}{c}{\multirow{3}{*}{---}} & \multirow{3}{*}{---} & 1.65 & 2.72 & 3.80 & 2.72 & 1.12 & \multirow{3}{*}{---} \\
& UniAD \cite{hu_planning-oriented_2023} \textcolor{gray}{\scriptsize [CVPR'23]}  & & & 1.26 & 2.22 & 3.06 & 2.18 & 1.13 & \\
& SparseDrive \cite{sun_sparsedrive_2025} \textcolor{gray}{\scriptsize [ICRA'25]} & & & 1.02 & 1.69 & 2.37 & 1.69 & 0.99 & \\

\midrule

% ROTATED GROUP 2: Coop V2X
\multirow{4}{*}{\rotatebox[origin=c]{90}{\scriptsize \textit{Coop V2X}}}
& V2VNet \cite{vedaldi_v2vnet_2020} \textcolor{gray}{\scriptsize [ECCV'20]}       & BEV Features & $8.19 \times 10^7$ & 1.96 & 2.37 & 3.41 & 2.58 & 0.88 & \multirow{4}{*}{---} \\
& CooperNaut \cite{cui_coopernaut_2022} \textcolor{gray}{\scriptsize [CVPR'22]}   & BEV Features & $8.19 \times 10^7$ & 2.69 & 4.07 & 5.50 & 4.09 & 1.42 & \\
& UniV2X \cite{yu_end--end_2025} \textcolor{gray}{\scriptsize [AAAI'25]}          & Queries \& Occ. & $8.09 \times 10^5$ & 1.45 & 2.19 & 3.04 & 2.23 & 0.25 & \\
& UniMM-V2X \cite{song_unimm-v2x_2025} \textcolor{gray}{\scriptsize [AAAI'26]}    & Queries \& Occ. & $9.32 \times 10^5$ & 0.78 & 1.63 & 2.05 & 1.49 & 0.12 & \\

\midrule

% ROTATED GROUP 3: Ours
\multirow{2}{*}{\rotatebox[origin=c]{90}{\scriptsize \textit{Ours}}}
& \textbf{\method}                                & \textbf{SDSM}                 & \textbf{1,408}  & \textbf{0.45} {\tiny $\pm 0.01$} & \textbf{0.83} {\tiny $\pm 0.01$} & \textbf{1.29} {\tiny $\pm 0.01$} & \textbf{0.86} {\tiny $\pm 0.01$} & \textbf{0.06} {\tiny $\pm 0.02$} & \textbf{88.33} {\tiny $\pm 0.23$} \\
% Grayed-out oracle row (colored cell-by-cell to prevent PDF rendering bugs with \pm):
& \textcolor{gray}{\method (w/ Map)}              & \textcolor{gray}{SDSM \& MAP} & \textcolor{gray}{25,792} & \textcolor{gray}{\textbf{0.45} {\tiny $\pm 0.00$}} & \textcolor{gray}{\textbf{0.83} {\tiny $\pm 0.00$}} & \textcolor{gray}{1.30 {\tiny $\pm 0.00$}} & \textcolor{gray}{\textbf{0.86} {\tiny $\pm 0.00$}} & \textcolor{gray}{\textbf{0.01} {\tiny $\pm 0.01$}} & \textcolor{gray}{\textbf{89.87} {\tiny $\pm 0.18$}} \\

\bottomrule
\end{tabular}
}
\vspace{-2em}
\end{table*}

\subsection{Two-Stage Training and Knowledge Transfer}

The decoupled sequence formulation of \method naturally supports a two-stage training pipeline, allowing the model to learn general driving kinematics from large-scale ego-centric data before efficiently adapting to additional V2X input. This training strategy directly addresses the data scarcity of real-world cooperative datasets by maximizing the utility of abundant single-vehicle driving logs.

\noindent\textbf{Ego-Centric Driving Pre-training:} In the first stage, \method is pre-trained on large-scale, single-agent datasets. During this phase, the generative planner establishes a foundational driving prior by learning to map front-view visual features, navigational commands, and optional topological map information to safe future trajectories. The model effectively captures core kinematic behaviors, such as lane following, intersection navigation, and basic obstacle avoidance, without requiring any infrastructure-provided context.

\noindent\textbf{V2X Cooperation Adaptation:} In the second stage, the model is fine-tuned on the target cooperative dataset (e.g., DAIR-V2X). Because \method processes inputs as a unified sequence, incorporating infrastructure data does not require modifying the model architecture; the encoded V2X tokens ($\bm{C}_{\text{v2x}}$) are simply appended to the single-agent context sequence $\bm{C}$. 

Throughout training, the visual backbone remains frozen to maintain a stable visual feature space. The generative planner leverages its pre-trained driving prior and adjusts its cross-attention distribution to attend to the newly appended V2X tokens. This structural continuity mitigates forgetting of the pre-trained driving prior and enables highly data-efficient transfer learning, allowing the model to adapt to cooperative settings using only a small fraction of the target dataset.

The primary supervision is the flow matching loss for future waypoint generation. For the optional map input consisting of multiple polylines, we employ an auxiliary map reconstruction head. A lightweight decoder reconstructs the map context sequence to predict lane-level polylines, thereby forcing the map message encoder to preserve all geometric information from the map. This auxiliary decoder is discarded during inference, incurring no deployment overhead.
\section{EXPERIMENTS}
\label{sec:experiments}

To evaluate the proposed \method framework, our experiments are designed to systematically validate its architectural advantages and the feasibility of practical deployment. First, we benchmark \method against state-of-the-art cooperative baselines, focusing on planning performance and communication efficiency. We then investigate whether our decoupled multi-modal token interface successfully bridges the V2X data scarcity gap via data-efficient transfer learning from a pre-trained ego-centric driving foundation model. Third, we validate the architectural generalizability of \method by comparing a Flow Matching planner against an Autoregressive (AR) planner. Finally, we test the system's robustness to real-world deployment challenges by injecting localization noise and conducting a V2X-augmented field test.

\subsection{Experimental Setup}
\label{sec:exp_setup}

\noindent\textbf{Datasets.}~We establish the foundational capability of single-agent end-to-end autonomous driving by conducting large-scale ego-centric pre-training on the \textit{navtrain} split of the nuPlan dataset~\cite{dauner_navsim_2024, caesar_nuplan_2022}. For cooperative fine-tuning and evaluation, we utilize the real-world \textit{DAIR-V2X-Seq}benchmark~\cite{yu_v2x-seq_2023}. We use the training and validation splits provided by UniV2X\cite{yu_end--end_2025} to ensure a fair comparison. This dataset features complex intersections and highly interactive traffic, providing a rigorous testbed for real-world V2X deployment. Because the nuPlan and DAIR-V2X-Seq datasets were collected in completely different regions, there is no data leakage between the validation sets in our setting. 

\noindent\textbf{Implementation Details.}~The visual perception module utilizes a frozen DINOv3-ViT-B/16 backbone~\cite{simeoni_dinov3_2025}. The shared hidden dimension across all tokenized modalities is set to $D=384$. The model has 130M parameters, of which 44M are trainable. For the SAE-compliant V2X infrastructure context, we limit the maximum token budget to 16 objects and 32 lanes per frame to guarantee low-latency processing. The objects from infrastructure-side perception are generated by a CenterPoint\cite{yin_center-based_2021} detector, which achieves an AP of $0.72$ for the vehicle category with $IoU~@~0.7$. The generative planner use 16 Transformer blocks and 8 attention heads, and is trained using Rectified Flow matching with an ODE integration horizon of $100$ steps. We truncate the trajectory generation to $20$ evaluation steps, striking an optimal balance between kinematic fidelity and execution efficiency. All models are trained using the AdamW optimizer. Pretraining was performed on a single NVIDIA A100-40GB for fewer than $40$ hours, and the V2X adaptation stage was performed on an  NVIDIA RTX 5090 for less than $10$ hours.

\noindent\textbf{Metrics.}~We evaluate open-loop trajectory planning quality using the Average $L_2$ Error (m) at $1.0$s, $2.0$s, and $3.0$s future horizons, following UniMMV2X~\cite{song_unimm-v2x_2025}. Communication efficiency is measured by Communication Cost (Bytes Per Second, BPS). Safety is evaluated via the open-loop Collision Rate (\%), calculated against ground-truth bounding boxes. To rigorously assess physical drivability, we employ the Predictive Driver Model Score (PDMS)~\cite{dauner_navsim_2024}. PDMS conducts an ego-closed-loop evaluation by treating the predicted trajectory as a reference command and evaluating a corresponding kinematically feasible updated trajectory. The final score evaluates the planning by multiplying strict binary safety criteria (At-fault Collision and Out of Road) with a weighted sum of continuous driving metrics (ego route progress, time-to-collision, and comfort). The missing routing and drivable-area information are inferred from the static map provided by the DAIR-V2X-Seq dataset. Unless otherwise specified, all results are obtained with a fixed seed, without cherry-picking.

\subsection{State-of-the-Art Cooperative Driving Benchmarking}
\label{sec:exp_main}

We compare \method against state-of-the-art single-agent baselines and V2X cooperative driving methods on the DAIR-V2X-Seq benchmark, following the same input settings as UniMMV2X~\cite{song_unimm-v2x_2025}. The quantitative results are summarized in Table~\ref{tab:main_results}.

\noindent\textbf{Projection-Free Superior Performance.}~As shown in Table~\ref{tab:main_results}, \method establishes a new state-of-the-art in end-to-end cooperative driving. Compared with previous methods, including the most recent UniMM-V2X~\cite{song_unimm-v2x_2025}, \method achieves the lowest Average $L_2$ Error ($0.86$m compared to $1.49$m) and reduces the Average Collision Rate to a near-perfect $0.06$\%, without ego-view perception-label supervision. Notably, as our method avoids the computation-intensive dense BEV feature processing and only executes token-level sparse computing, \method only requires \textbf{550 MB GPU memory} during inference, which is less than 8.5\% of the previous SoTA (UniMM-V2X~\cite{song_unimm-v2x_2025} requires $6483$ MB, as reported.) Due to the stochastic nature of the Flow Matching Transformer, we run the experiments with $10$ different random seeds and report the mean performance and standard deviation.

\noindent\textbf{Ultra-Low Bandwidth Consumption.}~A major deployment bottleneck for current cooperative driving is the communication bandwidth required to transmit features, and the mismatch between model-specific features and SAE-standard V2X messages. Recent query-based methods compress the heavy BEV feature payload into occupancy maps and queries, but they still consume $8.09 \times 10^5$ BPS. By seamlessly integrating SAE-standard Sensor Data Sharing Messages (SDSM) as independent conditioning tokens, \method reduces communication bandwidth requirements to \textbf{$\mathbf{1408}$ BPS}, a reduction of over $574 \times$ compared to previous methods under the same message rate assumption\footnote{The communication volume is calculated as $16~\text{objects}\times 44~\text{Bytes} \times 2~\text{FPS} = 1408~\text{BPS}$. An empirical communication volume, measured by the average size of the parsed JSON object, is 4231 BPS, corresponding to a $191\times$ reduction.}.

\begin{table}[t]
\centering
\caption{Metric based on Predictive Driver Model Scores (PDMS) for \method. Scores listed are the average of 5 runs with different seeds.}
\label{tab:pdms_detailed}
\resizebox{0.75\columnwidth}{!}{
\begin{tabular}{@{} l c c @{}}
\toprule
\textbf{Metric} & \textbf{\method} & \textbf{\method (w/ Map)} \\
\midrule
Comfort             & 99.10 & \textbf{99.74} \textcolor{green!50!black}{\scriptsize (+0.64)} \\
Drivable Area       & \textbf{94.92} & 94.78 \textcolor{red!70!black}{\scriptsize (-0.14)} \\
%Driving Direction   & 78.74 & \textbf{80.03} \textcolor{green!50!black}{\scriptsize (+1.29)} \\
Ego Progress        & 84.46 & \textbf{85.32} \textcolor{green!50!black}{\scriptsize (+0.86)} \\
No-Fault Collision  & 97.47 & \textbf{99.08} \textcolor{green!50!black}{\scriptsize (+1.61)} \\
Time-to-Collision   & 95.79 & \textbf{98.00} \textcolor{green!50!black}{\scriptsize (+2.21)} \\
\midrule
\textbf{Total PDMS} & 88.33 & \textbf{89.87} \textcolor{green!50!black}{\textbf{\scriptsize (+1.54)}} \\
\bottomrule
\end{tabular}
}
\vspace{-1.5em}
\end{table}

\noindent\textbf{Impact of Supplemented Static Map.}~When optionally supplemented with standard MAP messages (the \textcolor{gray}{gray} row in Table \ref{tab:main_results}), the $L_2$ error remains highly competitive, while the average collision rate further drops from $0.06\%$ to $0.01\%$. As the open-loop metrics become saturated, we compare the two variants using PDMS, as shown in Table~\ref{tab:pdms_detailed}. Safety-related metrics, including No-Fault Collision and Time-to-Collision, achieve the largest improvements, resulting in an overall PDMS improvement of $1.54$.

This indicates that explicit topological information from the map message serves as a structural hint that further refines the generation of continuous displacements over extended horizons. Although this information is not included in previously benchmarked methods, we believe that, in the context of vehicle-to-infrastructure (V2I) cooperation, receiving a vectorized map from roadside units is feasible and compliant with existing standards. Moreover, even with MAP messages included, the total bandwidth ($25792$ BPS) remains significantly lower than feature-based sharing paradigms.

\subsection{Data-Efficient Knowledge Transfer}
\label{sec:exp_transfer}

\begin{table}[tb]
    \centering
    \caption{Data-Efficient Adaptation on DAIR-V2X-Seq.}
    \label{tab:transfer}
    \resizebox{\columnwidth}{!}{
    \begin{tabular}{@{}lc|cc@{}}
    \toprule
    \textbf{Pre-training} & \textbf{Target V2X Data} & \textbf{Avg. L2 (m) $\downarrow$} & \textbf{Avg. PDMS $\uparrow$} \\
    \midrule
    From Scratch & 100\% & 1.5590 & 82.72 \\
    \midrule
    \multirow{5}{*}{\textit{navtrain} from nuPlan } 
    & 1\%   & 0.8894 \textcolor{green!50!black}{(-43.0\%)} & 88.04 \textcolor{green!50!black}{(+6.4\%)} \\
    & 10\%  & 0.8618 & 88.08 \\
    & 20\%  & 0.8585 & 88.20 \\
    & 50\%  & 0.8614 & 88.47 \\
    & 100\% & 0.8529 & 88.39 \\
    \bottomrule
    \end{tabular}
    }
    \vspace{-1em}
\end{table}

To validate the claim that the domain-stable sequence interface mitigates data scarcity, we evaluate the model fine-tuned on different split sizes. As detailed in Table~\ref{tab:transfer}, the limited size of the V2X dataset constrains the model's ability to learn basic planning capabilities. With our design of a stable context feature space, the generative foundation planner can be adapted to the new dataset using just 1\% of the V2X data. Increased V2X data scale improves the model's utilization of V2X messages. We owe the significant performance gain from pretraining to our design of unified context sequences, since existing methods also use pretrained BEVFormer\cite{li_bevformer_2025} as a starting point but do not achieve similar performance.

\subsection{System Ablations and Scalability}
\label{sec:exp_ablations}

\noindent\textbf{Generalizability to Different Generative Models.} 
To demonstrate that our framework is architecture-agnostic, we replace the cross-attention-conditioned Flow Transformer backbone with a prefix-conditioned Autoregressive (AR) backbone.
\begin{table}[tb]
    \centering
    \caption{Adaptation of Different Generative Models.}
    \label{tab:ar_vs_flow}
    \resizebox{\columnwidth}{!}{
    \begin{tabular}{@{}ll|cc@{}}
    \toprule
    \textbf{Generative Planner} & \textbf{Setting} & \textbf{Avg. L2 (m) $\downarrow$} & \textbf{Avg. PDMS $\uparrow$} \\
    \midrule
    \multirow{2}{*}{Autoregressive} 
    & From Scratch & 1.5420 & 76.55 \\
    & Pre-trained & 1.3499 \textcolor{green!50!black}{\scriptsize (-12.5\%)} & 80.31 \textcolor{green!50!black}{\scriptsize (+4.9\%)} \\
    \midrule
    \multirow{2}{*}{Flow Matching} 
    & From Scratch & 1.5590 & 82.72 \\
    & Pre-trained & 0.8529 \textcolor{green!50!black}{\scriptsize (-45.3\%)} & 88.39 \textcolor{green!50!black}{\scriptsize (+6.9\%)} \\
    \bottomrule
    \end{tabular}
    }
    \vspace{-1.5em}
\end{table}
As shown in Table \ref{tab:ar_vs_flow}, the AR backbone also demonstrates strong knowledge transfer capability across datasets. However, the Flow Transformer maintains a distinct advantage due to non-causal smoothing and significantly lower computational complexity via cross-attention injection.

\begin{figure}[tb]
  \centering
  \begin{subfigure}[t]{0.5\columnwidth}
    \centering
    \includegraphics[width=\linewidth]{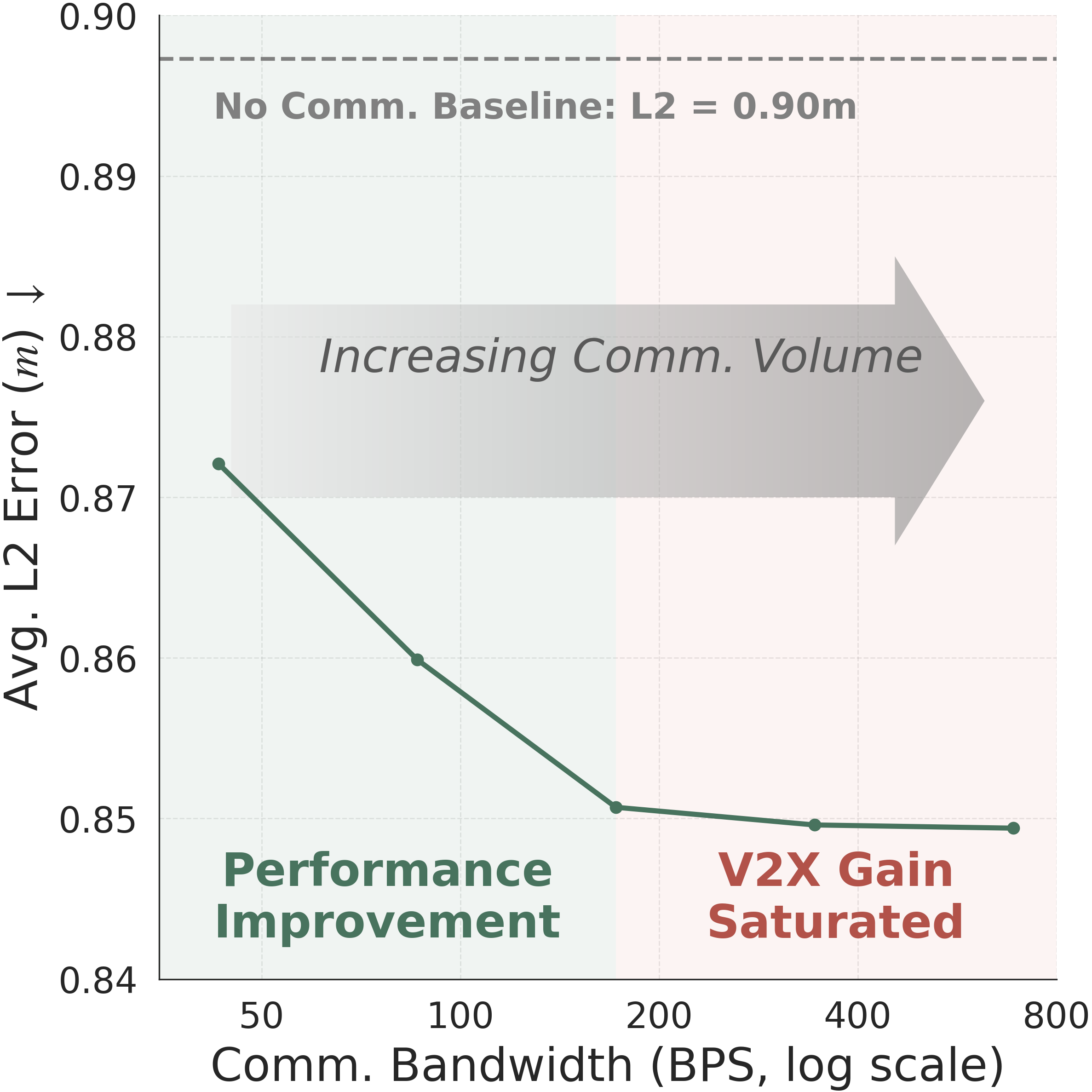}
    \vspace{-10pt}
    \caption{\footnotesize Communication Resource.}
    \label{fig:tradeoffs:comm}
  \end{subfigure}\hfill
  \begin{subfigure}[t]{0.5\columnwidth}
    \centering
    \includegraphics[width=\linewidth]{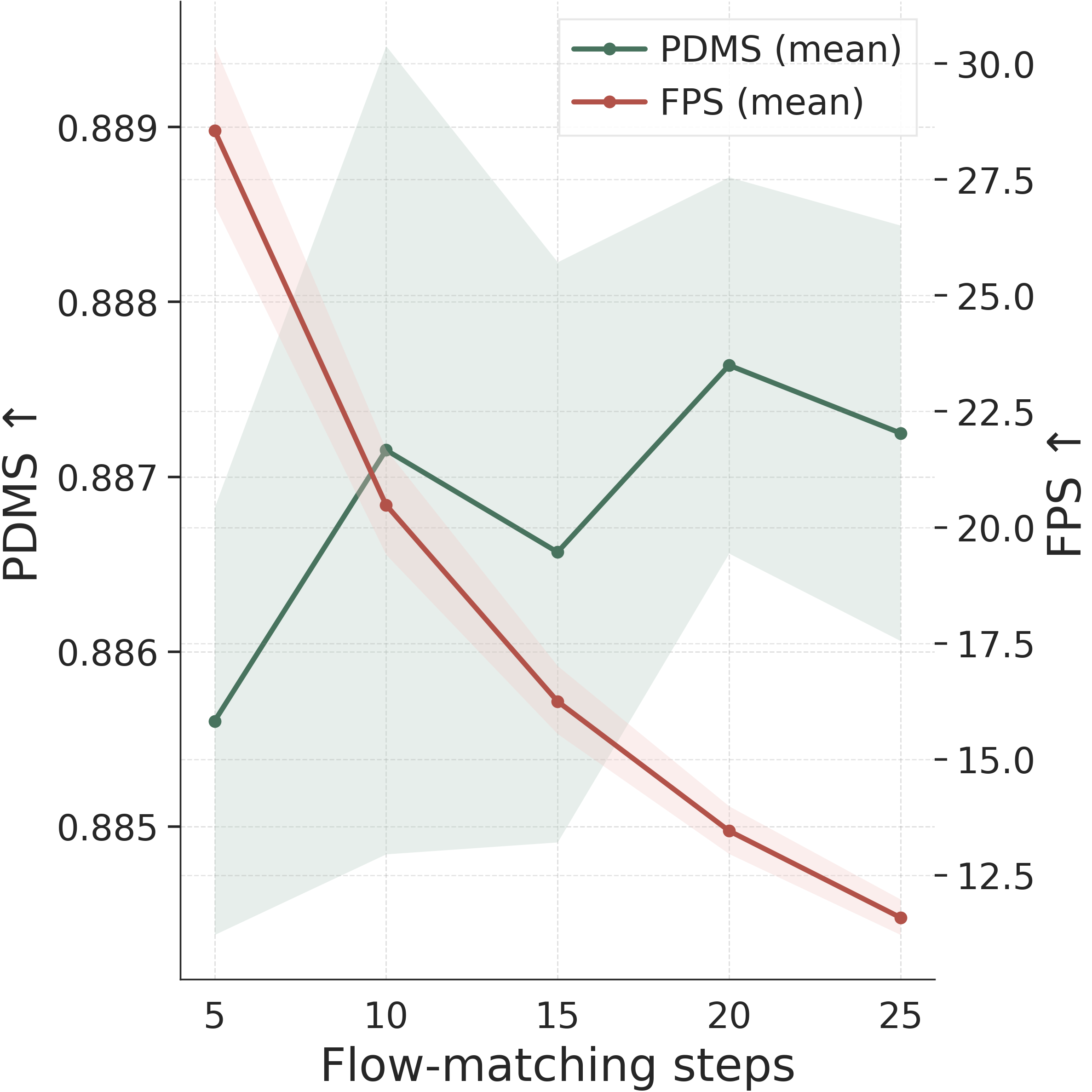}
    \vspace{-10pt}
    \caption{\footnotesize Computation Resource.}
    \label{fig:tradeoffs:steps}
  \end{subfigure}
  \caption{Performance-resource trade-offs.}
  \label{fig:tradeoffs}
  \vspace{-1em}
\end{figure}

\noindent\textbf{Communication \& Computation Resources Analysis.} 
Edge devices face strict constraints. In Fig. \ref{fig:tradeoffs:comm} (Left), we analyze the trade-off between communication bandwidth (scaling the token budget $M$) and planning accuracy. Though the overall average improvement of the L2 error is only 5.6\% due to the superior single-agent performance, we noticed a $\mathbf{11.9\%}$ improvement from no communication to abundant communication budget on a tough subset. We also tested the inference performance on an edge device (NVIDIA Jetson Thor) without quantization. In Fig. \ref{fig:tradeoffs:steps}, we demonstrate the flexible inference-time computation uniquely enabled by flow matching. An acceleration to a maximum of 29 FPS can be achieved without sacrificing much planning performance.

\subsection{System Robustness \& V2X-Augmented Field Testing}
\label{sec:exp_field}

While open-loop metrics evaluate the method using standardized evaluation protocols, cooperative planning safety depends heavily on robustness to real-world noisy inputs.

\begin{figure}[t]
    \centering
    \includegraphics[width=1\linewidth]{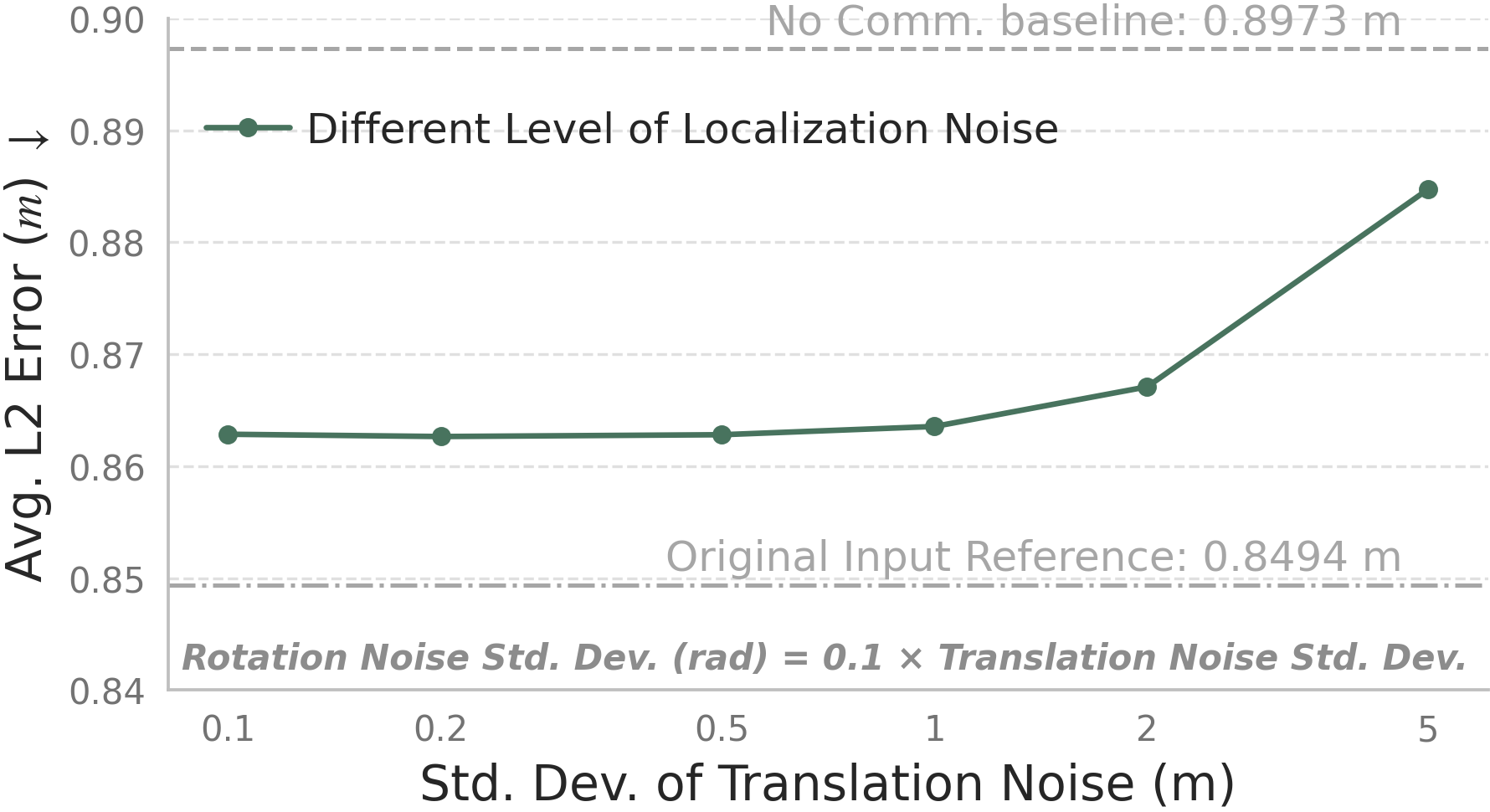}
    \caption{Robustness to ego localization error.}
    \label{fig:loc_error}
    \vspace{-1.5em}
\end{figure}

\noindent\textbf{Robustness to Input Degradation.}
We evaluate 3 deployment degradations: ego-localization perturbations, V2X latency, and infrastructure perception degradation.
In the \method framework, although we remove the need for precise calibration between ego-vehicle and infrastructure sensors by bypassing rigid spatial fusions, it still requires a reliable ego-vehicle pose to interpret object coordinates from GNSS. To evaluate the system's robustness to localization errors, we inject varying levels of noise into the relative translation and rotation matrices, ranging from a mild perturbation of $0.1~m/0.01~rad$ to an extreme perturbation of $5.0~m/0.5~rad$. Note that the noise is injected on top of the inherent noise of the real-world dataset. Results are shown in Figure \ref{fig:loc_error}, showing that \method maintains high robustness with a localization noise below $1.0~m/0.1~rad$. We also test the system's robustness towards latency and the quality of V2X messages: the average L2 error at an injected $500$~ms latency is $0.853$~m, and the average L2 error at a 30\% object dropout and a $0.5$~m detection error is $0.869$~m, both within a small margin of the original dataset setting.

% \begin{table}[h]
% \centering
% \resizebox{\columnwidth}{!}{
% \begin{tabular}{l|ccccccc}
% \toprule
% Metric & No Injected Noise & 0.1~m/0.01~rad & 0.2~m/0.02~rad & 0.5~m/0.05~rad & 1.0~m/0.10~rad & 2.0~m/0.20~rad & 5.0~m/0.50~rad \\
% \midrule
% Avg. L2 (m) & 0.849 & 0.863 & 0.863 & 0.863 & 0.864 & 0.867 & 0.885 \\
% \bottomrule
% \end{tabular}
% }
% \caption{Avg. L2 error across localization noise levels.}
% \label{tab:localization_noise}
% \vspace{-10pt}
% \end{table}

% --- REQ 2: FIELD TESTING (2-Column / Full Page Width, 4-in-a-row) ---

\noindent\textbf{V2X-Augmented Field Testing.} 
Finally, we deploy \method onto a real-world automated vehicle. We collect a single-vehicle dataset and augment it with virtual V2X messages and corresponding optimal waypoints. The augmented V2X messages have a simulated latency of $100$~ms, comprising the typical latency of roadside perception systems and the C-V2X sidelink communication latency. The dataset is collected at a parking lot. We consider it more challenging than the previously benchmarked dataset, as it includes severe occlusions, unclear lane markings, and less predictable agent movements than in typical urban scenarios. We split the data into 346 \textit{train} and 228 \textit{val} samples, strictly by isolated route, to minimize correlation between the two sets. Adapted from the model pretrained on \textit{navsim} and fine-tuned on DAIR-V2X-Seq, \method achieves an average L2 error of $1.14$~m on the validation set of our augmented dataset.

\begin{figure}[t]
    \centering
    \includegraphics[width=\columnwidth]{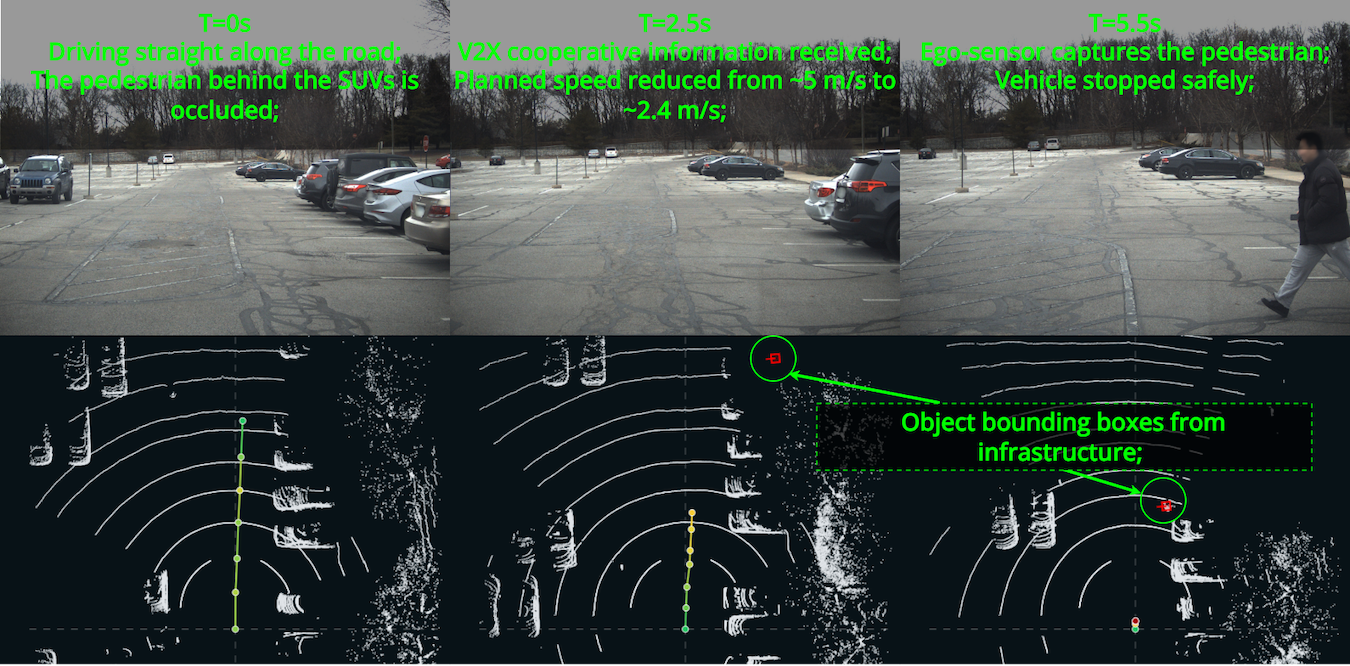}
    \caption{Real-world demonstration of V2X cooperative driving. The LiDAR point cloud is only used for visualization.}
    \label{fig:deployment}
    \vspace{-1.5em}
\end{figure}

As shown in the sequential rollout in Fig. \ref{fig:deployment}, we validate the system against a typical occlusion event: a pedestrian appears from behind a vehicle on the road. \method actively reacts to the pedestrian's existence and decelerates in advance, avoiding a hard brake or a potential crash.
\vspace{-0.5em}

\section{Conclusion}
In this paper, we presented \method, a scalable and highly efficient generative foundation planner that reshapes the paradigm of end-to-end cooperative driving. Departing from the conventional approach of computationally heavy and rigid spatial feature fusion, \method reformulates multi-agent cooperation as a multi-source conditional sequence generation task. By processing V2X messages and ego-centric visual features as independent context tokens via cross-attention injection, our framework successfully bypasses the spatial alignment constraints and memory bottlenecks of dense 3D representations. Extensive evaluations and robustness validation on the DAIR-V2X-Seq benchmark, our own augmented real-world dataset, and field testing, demonstrate that \method establishes a new state-of-the-art in planning safety and accuracy. More importantly, it operates at up to 29 FPS in real time, while consuming less than 1\% of the communication bandwidth and requiring 8.5\% of the inference memory, compared with existing methods. Furthermore, our unified context sequence effectively bridges the gap in cooperative data scarcity, enabling the model to adapt a pre-trained single-agent driving model to V2X scenarios using less than 10\% of the full target dataset.

\noindent\textbf{Limitations and Future Work.}~The imitation learning bottleneck is the main limitation of the benchmarked performance. While DAIR-V2X-Seq and its benchmark create a gap between ego-sensor-based planning and expert planning (human driver) by enabling only a front-view camera, fully V2X-assisted driving can only be achieved by an agent with complete V2X input. Future work will explore improving the method beyond human experts and scaling up the safety-critical testing from the current small-scale augmented dataset to large-scale closed-loop validation.

\vspace{-0.5em}

%\addtolength{\textheight}{-5cm}   % This command serves to balance the column lengths

%%%%%%%%%%%%%%%%%%%%%%%%%%%%%%%%%%%%%%%%%%%%%%%%%%%%%%%%%%%%%%%%%%%%%%%%%%%%%%%%

%%%%%%%%%%%%%%%%%%%%%%%%%%%%%%%%%%%%%%%%%%%%%%%%%%%%%%%%%%%%%%%%%%%%%%%%%%%%%%%%

%%%%%%%%%%%%%%%%%%%%%%%%%%%%%%%%%%%%%%%%%%%%%%%%%%%%%%%%%%%%%%%%%%%%%%%%%%%%%%%%
\bibliographystyle{IEEEtran}
\bibliography{references}

\end{document}